\documentclass[11pt,letterpaper]{article}
\usepackage[nohyperref]{emnlp2017}
\usepackage{times}
\usepackage{latexsym}
\usepackage{amsmath}
\usepackage{relsize}
\usepackage[utf8]{inputenc}
\usepackage{todonotes}
\usepackage{booktabs}

\emnlpfinalcopy

\def\emnlppaperid{***}

\title{{CUNI} System for the {WMT17} Multimodal Traslation Task}

\author{Jindřich Helcl \and Jindřich Libovický\\
  Charles University, Faculty of Mathematics and Physics \\
  Institute of Formal and Applied Linguistics \\
  Malostransk\' e n\' am\v est\' i 25, 118 00 Prague, Czech Republic \\
  {\tt \{helcl, libovicky\}@ufal.mff.cuni.cz}}

\date{}

\begin{document}

\maketitle

\begin{abstract}
  In this paper, we describe our submissions to the WMT17 Multimodal
  Translation Task. For Task~1 (multimodal translation), our best scoring
  system is a purely textual neural translation  of the source image caption to
  the target language. The main feature of the system is the use of additional
  data that was acquired by selecting similar sentences from parallel corpora
  and by data synthesis with back-translation. For Task~2 (cross-lingual image
  captioning), our best submitted system generates an English caption which is
  then translated by the best system used in Task~1. We also present negative
  results, which are based on ideas that we believe have potential of making
  improvements, but did not prove to be useful in our particular setup.
\end{abstract}

% ===========================================================================================
\section{Introduction}
% ===========================================================================================

Recent advances in deep learning allowed inferring distributed vector
representations of both textual and visual data. In models combining text and
vision modalities, this representation can be used as a shared data type.
Unlike the classical natural language processing tasks where everything happens
within one language or across languages, multimodality tackles how the language
entities relate to the extra-lingual reality. One of these tasks is multimodal
translation whose goal is using cross-lingual information in automatic image
captioning.

In this system-description paper, we describe our submission to the WMT17
Multimodal Translation Task. In particular, we discuss the effect of mining
additional training data and usability of advanced attention strategies. We
report our results on both the 2016 and 2017 test sets and discuss efficiency
of tested approaches.

The rest of the paper is organized as follows. Section~\ref{sec:task}
introduces the tasks we handle in this paper and the datasets that were
provided to the task. Section~\ref{sec:related} summarizes the
state-of-the-art methods applied to the task. In Section~\ref{sec:models}, we
describe our models and the results we have achieved.
Section~\ref{sec:negative} presents the negative results and
Section~\ref{sec:conclusion} concludes the paper.

% ===========================================================================================
\section{Task and Dataset Description}
\label{sec:task}
% ===========================================================================================

The challenge of the WMT Multimodal Translation Task is to exploit
cross-lingual information in automatic image caption generation. The
state-of-the-art models in both machine translation and automatic image caption
generation use similar architectures for generating the target sentence. The
simplicity with which we can combine the learned representations of various
inputs in a single deep learning model inevitably leads to a question whether
combining the modalities can lead to some interesting results. In the shared
task, this is explored in two subtasks with different roles of visual and
textual modalities.

In the multimodal translation task (Task~1), the input of the model is an image
and its caption in English. The system then should output a German or French
translation of the caption. The system output is evaluated using the
METEOR~\citep{denkowski2011meteor} and BLEU~\citep{papineni2002bleu} scores
computed against a single reference sentence. The question this task tries to
answer is whether and how is it possible to use visual information to
disambiguate the translation.

In the cross-lingual captioning task (Task~2), the input to the model at
test-time is the image alone. However, additionally to the image, the model is
supplied with the English (source) caption during training. The evaluation
method differs from Task~1 in using five reference captions instead of a single
one. In Task~2, German is the only target language. The motivation of Task~2 is
to explore ways of easily creating an image captioning system in a new language
once we have an existing system for another language, assuming that the
information transfer is less complex across languages than between visual and
textual modalities.

% -------------------------------------------------------------------------------------------
\subsection{Data}
% -------------------------------------------------------------------------------------------

\begin{table} 
 
\begin{center}
\begin{tabular}{l|c|c|c}
& en & de & fr \\ \hline \hline
Train.\ sentences & \multicolumn{3}{c}{29,000} \\ \hline
Train.\ tokens & 378k & 361k & 410k \\
Avg. \# tokens & 13.0 & 12.4 & 14.1 \\
\# tokens range & 4--40 & 2--44 & 4--55 \\ \hline
Val.\ sentences & \multicolumn{3}{c}{1,014} \\ \hline
Val.\ tokens & 13k & 13k & 14k \\
Avg.\ \# tokens & 13.1 & 12.7 & 14.2 \\
\# tokens range & 4--30 & 3--33 & 5--36 \\ \hline
OOV rate & 1.28\% & 3.09\% & 1.20\%
\end{tabular}
\end{center}

\caption{Multi30k statistics on training and validation data -- total number of
tokens, average number of tokens per sentence, and the sizes of the shortest
and the longest sentence.}\label{tab:data}

\end{table}

The participants were provided with the Multi30k
dataset~\citep{elliott2016multi30k} -- a~multilingual extension of Flickr30k
dataset~\citep{plummer2017flickr30k} -- for both training and evaluation of
their models.

The data consists of 31,014 images. In Flickr30k, each image is described with
five independently acquired captions in English. Images in the Multi30k dataset
are enriched with five crowd-sourced German captions. Additionally, a single
German translation of one of the English captions was added for each image.

The dataset is split into training, validation, and test sets of 29,000, 1,014,
and 1,000 instances respectively. The statistics on the training and validation
part are tabulated in Table~\ref{tab:data}.

For the 2017 round of the competition, an additional French translation was
included for Task~1 and new test sets have been developed. Two test sets were
provided for Task~1: The first one consists of 1,000 instances and is similar
to the test set used in the previous round of the competition (and to the
training and validation data). The second one consists of images, captions,
and their translations taken from the MSCOCO image captioning
dataset~\citep{tsung2014mscoco}. A new single test set containing 1,071 images
with five reference captions was added for Task~2.

The style and structure of the reference sentences in the Flickr- and
MSCOCO-based test sets differs. Most of the sentences in the Multi30k dataset
have a similar structure with a relatively simple subject, an active verb in
present tense, simple object, and location information (e.g., ``\emph{Two dogs
are running on a beach.}''). Contrastingly, the captions in the MSCOCO dataset
are less formal and capture the annotator's uncertainty about the image content
(e.g., ``\emph{I don't know, it looks like a lemon.}'').

% ===========================================================================================
\section{Related Work}
\label{sec:related}
% ===========================================================================================

Several promising neural architectures for multimodal translation task have
been introduced since the first competition in 2016.

In our last year's submission~\citep{libovicky2016cuni}, we employed a neural
system that combined multiple inputs -- the image, the source caption and an
SMT-generated caption. We used the attention mechanism over the textual
sequences and concatenated the context vectors in each decoder step.

The overall results of the WMT16 multimodal translation task did not prove the
visual features to be particularly
useful~\citep{specia2016shared,caglayan2016multimodality}.

To our knowledge, \citet{huang2016attention} were the first who showed
an improvement over a textual-only neural system with model
utilizing distributed features explicit object recognition. \citet{calixto2017incorporating}
improved state of the art using a model initializing the decoder state with the
image vector, while maintaining the rest of the neural architecture unchanged.
Promising results were also shown by \citet{delbrouck2017multimodal} who made a
small improvement using bilinear pooling.

\citet{elliot2017imagination} brought further improvements by introducing the
``imagination'' component to the neural network architecture. Given the source
sentence, the network is trained to output the target sentence jointly with
predicting the image vector. The model uses the visual information only as a
regularization and thus is able to use additional parallel data without
accompanying images.

\begin{figure*}[t]
\begin{center}
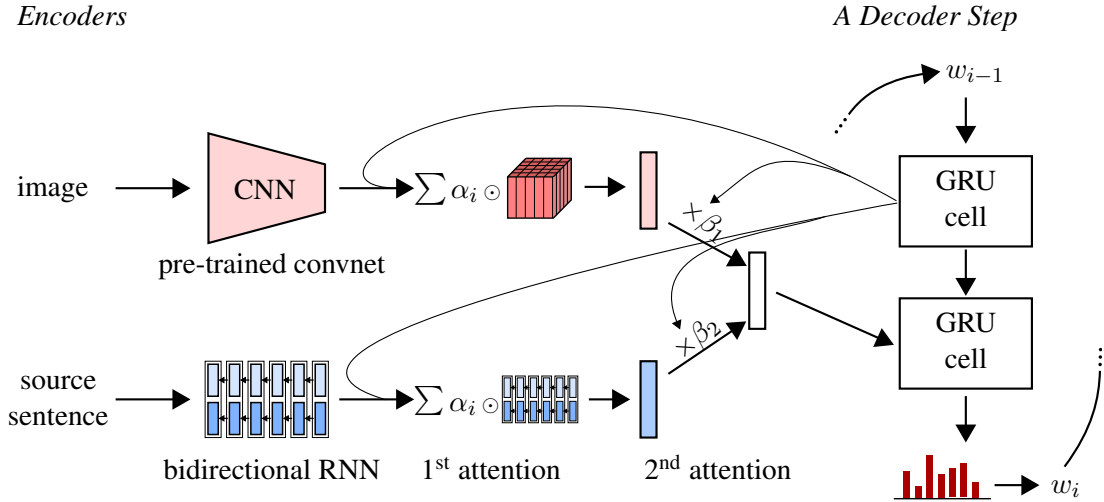
\end{center}

\caption{An overall picture of the multimodal model using hierarchical
attention combination on the input. Here, $\alpha$ and $\beta$ are normalized
coefficients computed by the attention models, $w_i$ is the $i$-th input to the
decoder.}\label{fig:hier}

\end{figure*}

% ===========================================================================================
\section{Experiments}
\label{sec:models}
% ===========================================================================================

All models are based on the encoder-decoder architecture with attention
mechanism~\cite{bahdanau2015neural} as implemented in Neural
Monkey~\citep{NeuralMonkey:2017}.\footnote{https://github.com/ufal/neuralmonkey}
The decoder uses conditional GRUs~\cite{firat2016cgru} with 500 hidden units
and word embeddings with dimension of 300. The target sentences are decoded
using beam search with beam size 10, and with exponentially weighted length
penalty~\citep{google2016bridging} with $\alpha$ parameter empirically
estimated as 1.5 for German and 1.0 for French. Because of the low OOV rate
(see Table~\ref{tab:data}), we used vocabularies of maximum 30,000 tokens and
we did not use sub-word units. The textual encoder is a bidirectional GRU
network with 500 units in each direction and word embeddings with dimension of
300. We use the last convolutional layer VGG-16
network~\citep{simonyan2014vgg} of dimensionality $14\times14\times512$ for
image processing. The model is optimized using the Adam optimizer
\citep{kingma2015adam} with learning rate $10^{-4}$ with early stopping based
on validation BLEU score.

% -------------------------------------------------------------------------------------------
\subsection{Task 1: Multimodal Translation}
% -------------------------------------------------------------------------------------------

We tested the following architectures with different datasets (see
Section~\ref{ssec:additional} for details):

\begin{itemize}
\item purely textual (disregarding the visual modality);
\item multimodal with context vector concatenation in the
    decoder~\citep{libovicky2016cuni};
\item multimodal with hierarchical attention
    combination~\citep{libovicky2017attention} -- context vectors are computed
    independently for each modality and then they are combined together using
    another attention mechanism as depicted in Figure~\ref{fig:hier}.
\end{itemize}

% -------------------------------------------------------------------------------------------
\subsection{Task 2: Cross-lingual Captioning}
% -------------------------------------------------------------------------------------------

We conducted two sets of experiments for this subtask. In both of them, we used
an attentive image captioning model~\citep{xu2015show} for the cross-lingual
captioning with the same decoder as for the first subtask.

The first idea we experimented with was using a multilingual decoder provided
with the image and a language identifier. Based on the identifier, the decoder
generates the caption either in English or in German. We speculated that the
information transfer from the visual to the language modality is the most
difficult part of the task and might be similar for both English and German.

The second approach we tried has two steps. First, we trained an English image
captioning system, for which we can use larger datasets. Second, we translated
the generated captions with the multimodal translation system from the first
subtask.

% -------------------------------------------------------------------------------------------
\subsection{Acquiring Additional Data}
\label{ssec:additional}
% -------------------------------------------------------------------------------------------

In order to improve the textual translation, we acquired additional data. We
used the following technique to select in-domain sentences from both parallel
and monolingual data.

We trained a neural character-level language model on the German sentences
available in the training part of the Multi30k dataset. We used a GRU network
with 512 hidden units and character embedding size of 128.

Using the language model, we selected 30,000 best-scoring German sentences from
the SDEWAC corpus~\citep{faass2013sdewac} which were both semantically and
structurally similar to the sentences in the Multi30k dataset.

% Finally, we back-translated these selected sentences along with the original sentences from Multi30k
% English and added them to the training data~\citep{sennrich2016backtranslation}.

We tried to use the language model to select sentence pairs also from parallel
data. By scoring the German part of several parallel corpora (EU
Bookshop~\citep{tiedemann2014billions}, News
Commentary~\citep{tiedemann2012parallel} and
CommonCrawl~\citep{smith2013dirt}), we were only able to retrieve a few
hundreds of in-domain sentences. For that reason we also included sentences
with lower scores which we filtered using the following rules: sentences must
have between 2 and 30 tokens, must be in the present tense, must not contain
non-standard punctuation, numbers of multiple digits, acronyms, or named
entities, and must have at most 15~\% OOV rate w.r.t. Multi30k training
vocabulary. We extracted additional 3,000 in-domain parallel sentences using
these rules. Examples of the additional data are given in
Table~\ref{tab:example}.

By applying the same approach on the French versions of the corpora, we were
pable to extract only few additional in-domain sentences. We thus
trained the English-to-French models in the constrained setup only.

Following \citet{calixto2017incorporating}, we
back-translated~\citep{sennrich2016backtranslation} the German captions from
the German side of the Multi30k dataset (i.e. 5+1 captions for each image), and
sentences retrieved from the SDEWAC corpus. We included these back-translated
sentence pairs as additional training data for the textual and multimodal
systems for Task~1. The back-translation system used the same architecture as
the textual systems and was trained on the Multi30k dataset only.
The additional parallel data and data from the SDEWAC corpus (denoted as
additional in Table~\ref{tab:task1}) were used only
for the text-only systems because they were not accompanied by images.

For Task~2, we also used the MSCOCO~\citep{tsung2014mscoco} dataset which
consists of 300,000 images with 5 English captions for each of them.

\begin{table}[t]

\makeatletter
\newcommand \Dotfill {\leavevmode \cleaders \hb@xt@ .80em{\hss $\cdot$\hss }\hfill \kern \z@}
\makeatother

\newcommand{\paraexl}[2]{#1\Dotfill \vspace*{-6pt}
\begin{flushright}\Dotfill \emph{#2} \end{flushright} \vspace*{-6pt}
}

\newcommand{\paraex}[2]{#1\Dotfill \emph{#2} \\}

\rule{\columnwidth}{0.4pt}
\begin{center} \vspace*{-6pt} \textbf{SDEWAC Corpus} (with back-translation) \end{center}%
\vspace{-14pt}
\rule{\columnwidth}{0.4pt}
\paraexl{zwei Männer unterhalten sich}{two men are talking to each other .}
%\paraexl{Menschen auf der Straße .}{People on the street .}
\paraexl{ein kleines Mädchen sitzt auf einer Schaukel .}{a little girl is sitting on a swing .}
%\paraexl{Eine junge Frau sitzt auf einer Bank und liest ein Buch .}{A young woman sits on a bench reading a book .}
\paraexl{eine Katze braucht Unterhaltung .}{a cat is having a discussion .}
\paraexl{dieser Knabe streichelt das Schlagzeug .}{this professional is petting the drums .}
\vspace{-6pt}
\rule{\columnwidth}{0.4pt}
\begin{center} \vspace*{-6pt} \textbf{Parallel Corpora}\end{center} 
\vspace{-14pt}\rule{\columnwidth}{0.4pt}
\paraex{Menschen bei der Arbeit}{People at work}
%\paraexl{Kinder und Jugendliche in der Stadt}{Children and young people in the urban environment}
\paraex{Männer und Frauen}{Men and women}
\paraex{Sicherheit bei der Arbeit}{Safety at work}
\paraexl{Personen in der Öffentlichkeit}{Members of the public}
%\paraexl{Frauen und Männer}{Women and men}

\caption{Examples of the collected additional training data.}
\label{tab:example}
\end{table}

% ===========================================================================================
\subsection{Results}
%\label{sec:experiments}
% ===========================================================================================

\begin{table*}[t]
\begin{center}
\begin{tabular}{lc||c|c|c||c|c}
 & & \multicolumn{3}{c||}{Task 1: en $\rightarrow$ de} & \multicolumn{2}{c}{Task 1: en $\rightarrow$ fr} \\ \cline{3-7}
 & & 2016 & Flickr & MSCOCO & Flickr & MSCOCO \\ \hline \hline

Baseline & C &
             --- & 19.3 / 41.9 & 18.7 / 37.6 & 44.3 / 63.1 & 35.1 / 55.8 \\ \hline

Textual & C &
     34.6 / 51.7 & 28.5 / 49.2 & 23.2 / 43.8 & {\bf 50.3} / 67.0 & {\bf 43.0} / {\bf 62.5} \\ 
Textual \small (+ Task2) & U & 
     36.6 / 53.0 & 28.5 / 45.7 & 24.1 / 40.7	  & --- & --- \\ 
Textual \small (+ additional) & U & 
     {\bf 36.8} / {\bf 53.1} & {\bf 31.1} / {\bf 51.0} & {\bf 26.6} / {\bf 46.0} & --- & --- \\ \hline

Multimodal {\small (concat.\ attn)} & C & 
     32.3 / 50.0 & 23.6 / 41.8 & 20.0 / 37.1 & 40.3 / 56.3 & 32.8 / 52.1 \\ 
Multimodal {\small (hier.\ attn.)} & C &
     31.9 / 49.4 & 25.8 / 47.1 & 22.4 / 42.7 & 49.9 / {\bf 67.2} & 42.9 / {\bf 62.5} \\ 
Multimodal {\small (concat.\ attn.)} & U &
     {\bf 36.0} / {\bf 52.1} & 26.3 / 43.9 & 23.3 / 39.8  & --- & --- \\ 
Multimodal {\small (hier.\ attn.)} & U &
     34.4 / 51.7 & {\bf 29.5} / {\bf 50.2} & {\bf 25.7} / {\bf 45.6} & --- & --- \\ \hline

Task~1 winner (LIUM-CVC) & C &
     --- & 33.4 / 54.0 & 28.7 / 48.9 & 55.9 / 72.1 & 45.9 / 65.9 
     
\end{tabular}
\end{center}

\caption{Results of Task 1 in BLEU / METEOR points. `C' denotes constrained
    configuration, `U' unconstrained, `2016' is the 2016 test set, `Flickr' and
    `MSCOCO' denote the 2017 test sets. The two unconstrained textual models
    differ in using the additional textual data, which was not used for the
training of the multimodal systems.}\label{tab:task1}

\end{table*}

In Task~1, our best performing system was the text-only system trained with
additional data. These were acquired both by the data selection method
described above and by back-translation. Results of all setups for Task~1 are
given in Table~\ref{tab:task1}.

Surprisingly, including the data for Task~2 to the training set decreased the 
METEOR score on both of the 2017 test sets. This might have been caused by domain
mismatch. However, in case of the additional parallel and SDEWAC data,
this problem was likely outweighed by the advantage of having more training data.

In case of multimodal systems, adding approximately the same amount of data
increased the performance more than in case of the text-only system. This
suggests, that with sufficient amount of data (which is a rather unrealistic
assumption), the multimodal system would eventually outperform the textual one.

The hierarchical attention combination brought major improvements over the
concatenation approach on the 2017 test sets. On the 2016 test set, 
the concatenation approach yielded better results, which can be considered
a somewhat strange result, given the similarity of the Flickr test sets.

The baseline system was Nematus~\citep{sennrich2017nematus} trained on the
textual part of Multi30k only. However, due to its low score, we suspect the
model was trained with suboptimal parameters because it is in principle a model
identical to our constrained textual submission.

In Task~2, none of the submitted systems outperformed the baseline which was a
captioning system~\citep{xu2015show} trained directly on the German captions in
the Multi30k dataset. The results of our systems on Task~2 are shown in
Table~\ref{tab:task2}.

\begin{table}
\begin{center}
\begin{tabular}{lc||c}
 &  & Task 2 \\ \hline \hline
Baseline                     & C &    {\bf 9.1} / {\bf 23.4} \\ \hline
Bilingual captioning         & C &    2.3 / 17.6 \\
en captioning + translation  & C &    4.2 / 22.1 \\
en captioning + translation  & U &    6.5 / 20.6 \\ \hline
other participant            & C &    {\bf 9.1} / 19.8
\end{tabular}
\end{center}
\caption{Results of Task 2 in BLEU / METEOR points.}
\label{tab:task2}
\end{table}

\begin{table}
\begin{center}
\begin{tabular}{l||c}
 & Flickr30k \\ \hline \hline
\citet{xu2015show}       & {\bf 19.1} / 18.5 \\ \hline
ours: Flickr30k          & 15.3 / {\bf 18.7} \\
ours: Flickr30k + MSCOCO & 17.9 / 16.6 \\
\end{tabular}
\end{center}
\caption{Results of the English image captioning systems on Flickr30k test set in
BLEU / METEOR points}
\label{tab:captioning}
\end{table}

For the English captioning, we trained two models. First one was trained on the
Flickr30k data only. In the second one, we included also the MSCOCO dataset.
Although the captioning system trained on more data achieved better performance
on the English side (Table~\ref{tab:captioning}), it led to extremely low
performance while plugged into our multimodal translation systems
(Table~\ref{tab:task2}, rows labeled ``en captioning + translation''). We
hypo\-the\-size this is caused by the different styles of the sentences in the
training datasets.

Our hypothesis about sharing information between the languages in a single
decoder was not confirmed in this setup and the experiments led to relatively
poor results.

Interestingly, our systems for Task~2 scored poorly in the BLEU score and 
relatively well in the METEOR score. We can attribute this to the fact that
unlike BLEU which puts more emphasis on precision, METEOR considers 
strongly also recall.

% ===========================================================================================
\section{Negative Results}
\label{sec:negative}
% ===========================================================================================

In addition to our submitted systems, we tried a number of techniques without
success. We describe these techniques since we believe it might be relevant
for future developments in the field, despite the current negative result.

% -------------------------------------------------------------------------------------------
\subsection{Beam Rescoring}
% -------------------------------------------------------------------------------------------

Similarly to \citet{lala2017unraveling}, our oracle experiments on the
validation data showed that rescoring of the decoded beam of width 100 has the
potential of improvement of up to 3 METEOR points. In the oracle experiment,
we always chose a sentence with the highest sentence-level BLEU score.
Motivated by this observation, we conducted several experiments with beam
rescoring.

%hoping that this may a better way of introducing the multimodal information to the purely
%textual translation than incorporating the image features in the model directly.

We trained a classifier predicting whether a given sentence is a suitable
caption for a given image. The classifier had one hidden layer with 300 units
and had two inputs: the last layer of the VGG-16 network processing the image,
and the last state of a bidirectional GRU network processing the text. We used
the same hyper-parameters for the bidirectional GRU network as we did for
the textual encoders in other experiments. Training data were taken from both parts
of the Multi30k dataset with negative examples randomly sampled from the
dataset, so the classes were represented equally. The classifier achieved
validation accuracy of 87\% for German and 74\% for French. During the
rescoring of the 100 hypotheses in the beam, we selected the one which had the
highest predicted probability of being the image's caption.

In other experiments, we tried to train a regression predicting the score of a
given output sentence. Unlike the previous experiment, we built the training
data from scored hypotheses from output beams obtained by translating the
training part of the Multi30k dataset. We tested two architectures: the first
one concatenates the terminal states of bidirectional GRU networks encoding the
source and hypothesis sentences and an image vector; the second performs an
attentive average pooling over hidden states of the RNNs and the image CNN
using the other encoders terminal states as queries and concatenates the
context vectors. The regression was estimating either the sentence-level BLEU
score~\citep{chen2014sbleu} or the chrF3 score~\citep{popovic2015chrf}.

Contrary to our expectations, all the rescoring techniques decreased the
performance by 2 METEOR points.

% -------------------------------------------------------------------------------------------
\subsection{Reinforcement Learning}
% -------------------------------------------------------------------------------------------

Another technique we tried without any success was self-critical sequence
training~\citep{rennie2016self}. This modification of the REINFORCE
algorithm~\citep{williams1992simple} for sequence-to-sequence learning uses the
reward of the training-time decoded sentence as the baseline. The systems were
pre-trained with the word-level cross-entropy objective and we hoped to
fine-tune the systems using the REINFORCE towards sentence-level BLEU score and
GLEU score~\citep{google2016bridging}.

It appeared to be difficult to find the right moment when the optimization
criterion should be switched and to find an optimal mixing factor of the
cross-entropy loss and REINFORCE loss. We hypothesize that a more complex
objective mixing strategy (like MIXER~\citep{ranzato2015mixer}) could lead to
better results than simple objective weighting.

% ===========================================================================================
\section{Conclusions}
\label{sec:conclusion}
% ===========================================================================================

In our submission to the 2017 Multimodal Task, we tested the advanced attention
combination strategies~\citep{libovicky2017attention} in a more challenging
context and  achieved competitive results compared to other submissions. We
explored ways of acquiring additional data for the task and tested two
promising techniques that did not bring any improvement to the system
performance.

% ===========================================================================================
\section*{Acknowledgments}

This research has been funded by the Czech Science Foundation grant no.
P103/12/G084, the EU grant no. H2020-ICT-2014-1-645452 (QT21), and Charles
University grant no. 52315/2014 and SVV project no. 260 453. 

This work has
been using language resources developed and/or stored and/or distributed by the
LINDAT-Clarin project of the Ministry of Education of the Czech Republic
(project LM2010013).

%\todo[inline]{Do camera ready verze napsat místo "Task 1 winner" citaci toho jejich system description papery - ale já ji neznám}
% SVV
% Jindruv GAUK
% QT 21 ??;	
% Multimodální GAČR

\bibliography{emnlp2017}
\bibliographystyle{emnlp_natbib}

\end{document}